\documentclass[runningheads]{llncs}

\usepackage{booktabs}
\usepackage{multirow}
\usepackage{xcolor}
\usepackage[T1]{fontenc}
\usepackage{graphicx}
\usepackage{float}
\usepackage{graphicx}
\usepackage{booktabs}

\begin{document}

\title{A Comparative Study of Adaptation Strategies for Time Series Foundation Models in Anomaly Detection}
\titlerunning{A Comparative Study of Adaptation Strategies for TSFMs in TSAD}

\author{
Miseon Park\inst{1}\and
Kijung Yoon\inst{1,2}
}

\authorrunning{M. Park et al.}

\institute{
Department of Electronic Engineering, Hanyang University, Seoul, Korea \and
Department of Artificial Intelligence, Hanyang University, Seoul, Korea \\
\email{\{misun9631,kiyoon\}@hanyang.ac.kr}
}

\maketitle 

\begin{abstract}
Time series anomaly detection is essential for the reliable operation of complex systems, but most existing methods require extensive task-specific training. We explore whether time series foundation models (TSFMs), pretrained on large heterogeneous data, can serve as universal backbones for anomaly detection. Through systematic experiments across multiple benchmarks, we compare zero-shot inference, full model adaptation, and parameter-efficient fine-tuning (PEFT) strategies. Our results demonstrate that TSFMs outperform task-specific baselines, achieving notable gains in AUC-PR and VUS-PR, particularly under severe class imbalance. Moreover, PEFT methods such as LoRA, OFT, and HRA not only reduce computational cost but also match or surpass full fine-tuning in most cases, indicating that TSFMs can be efficiently adapted for anomaly detection, even when pretrained for forecasting. These findings position TSFMs as promising general-purpose models for scalable and efficient time series anomaly detection.

\end{abstract}

\section{Introduction}
Time series anomaly detection (TSAD) is a fundamental problem in data-driven monitoring systems, playing a vital role in maintaining the reliability, security, and safety of real-world applications ranging from industrial process control to financial surveillance \cite{blazquez2021review,chandola2009anomaly}. The goal of TSAD is to identify abnormal behaviors in sequential data that deviate significantly from expected temporal patterns. Over the years, two dominant methodological paradigms have emerged to tackle this problem: forecasting-based and reconstruction-based approaches. Forecasting-based methods learn temporal dependencies from historical observations to predict future values, and anomalies are identified when observed data exhibit large deviations from these predictions \cite{deng2021graph,chen2021learning,wu2023timesnet}. Owing to their causal nature, these methods are particularly well suited for online detection scenarios, where only past information is available at each time step. Typical implementations rely on predictive models such as recurrent neural networks \cite{ergen2019unsupervised,hundman2018detecting}, Transformers \cite{xu2022anomaly,tuli2022tranad}, or temporal graph neural networks \cite{zhao2020multivariate,deng2021graph}, with prediction errors serving as anomaly scores. In contrast, reconstruction-based approaches aim to encode and reconstruct the input time series, treating large reconstruction errors as indicators of anomalous behavior \cite{park2018multimodal,li2019mad,xu2022anomaly}. These methods are generally applied in retrospective settings where full sequences are available, under the assumption that models trained predominantly on normal data cannot faithfully reconstruct rare or abnormal patterns. Despite their demonstrated effectiveness, both paradigms typically rely on substantial task-specific training and careful tuning, which limits their adaptability across diverse real-world settings.

Recently, time series foundation models (TSFMs) have emerged as a general-purpose paradigm for time series representation learning and forecasting \cite{yue2022ts2vec,jin2024position,ansari2024chronos,woo2024unified,talukder2024totem}. Importantly, the development of TSFMs is not driven by anomaly detection in isolation, but by the broader objective of learning transferable temporal representations from large-scale and heterogeneous time series data. Inspired by foundation models in natural language processing, TSFMs are pretrained to capture universal temporal structures that can support a wide range of downstream tasks. Among these tasks, forecasting-based anomaly detection naturally benefits from the strong predictive capabilities of TSFMs, as anomalies can be identified through deviations between forecasts and observations. In practice, TSFMs generate multi-step predictions conditioned only on past observations, enabling their direct use in online detection settings. Moreover, their strong generalization ability allows for zero-shot anomaly detection, where anomalies can be flagged without task-specific retraining. Prior works have shown that models such as TimeGPT \cite{garza2023timegpt}, Chronos \cite{ansari2024chronos}, Moirai \cite{woo2024unified}, and Time-MoE \cite{shi2025timemoe} achieve competitive forecasting performance on previously unseen datasets, highlighting their potential as broadly applicable forecasting backbones.

In parallel with the rise of foundation models, parameter-efficient fine-tuning (PEFT) has been proposed as a general strategy for adapting large pretrained models to downstream tasks. Similar to TSFMs, PEFT methods were not originally developed for time-series forecasting or anomaly detection, but rather to address the broader challenge of efficiently specializing large-scale models without updating all parameters. Techniques such as LoRA \cite{hu2022lora}, IA$^3$ \cite{liu2022fewshot}, OFT \cite{qiu2023controlling}, and HRA \cite{yuan2024bridging} enable adaptation by modifying only a small subset of parameters or by inserting lightweight modules. When applied to TSFMs, these approaches provide a practical means of adjusting forecasting behavior to domain-specific characteristics, which can in turn improve anomaly detection performance. Crucially, PEFT allows practitioners to exploit the generalization ability of pretrained TSFMs while keeping computational and memory costs manageable.

In this work, we study how general-purpose time series foundation models and parameter-efficient fine-tuning techniques can be \textit{repurposed} for time series anomaly detection. Our hypothesis is that TSFMs pretrained on large and diverse datasets encode transferable temporal knowledge that serves as a strong prior for distinguishing normal and abnormal behaviors. By applying lightweight PEFT adaptations, this prior can be efficiently aligned with specific anomaly detection settings without the expense of full fine-tuning. Although TSFMs and PEFT have been extensively investigated in their respective contexts, their combined use for TSAD remains underexplored. To address this gap, we conduct a systematic empirical evaluation across multiple TSAD benchmarks, comparing zero-shot inference, full fine-tuning, and PEFT-based adaptation. We further analyze the impact of evaluation metrics and PEFT hyperparameters on detection performance. Our contributions are summarized as follows:
\begin{itemize}
    \item We present a systematic empirical study on repurposing time series foundation models for time series anomaly detection, comparing zero-shot inference, full fine-tuning, and parameter-efficient fine-tuning strategies across multiple architectures and benchmarks.
    \item We demonstrate that adapted TSFMs consistently outperform task-specific baselines, with particularly strong gains on precision-oriented and VUS-based metrics under severe class imbalance, highlighting the importance of evaluation protocol and adaptation strategy in TSAD.
    \item We show that PEFT methods such as LoRA, OFT, and HRA often match or surpass full fine-tuning while updating only a small fraction of model parameters, enabling efficient adaptation of both dense and mixture-of-experts TSFMs without training models from scratch.
\end{itemize}

\section{Related Works}
\subsection{Time Series Foundation Models}
Time series foundation models have recently emerged as a general-purpose paradigm for learning transferable temporal representations, with forecasting as their primary pretraining objective. Broadly, existing TSFMs can be categorized by how they leverage large-scale data and architectures originally developed for other domains. Models such as Moirai \cite{woo2024unified} are trained directly on massive and heterogeneous time series corpora, enabling them to capture universal temporal patterns across domains. In contrast, Chronos \cite{ansari2024chronos} adapts large language models (LLMs) to time series forecasting by tokenizing numerical values and applying pretrained transformer-based language models such as T5 \cite{raffel2020exploring} and GPT-2 \cite{radford2019language}. Time-MoE \cite{shi2025timemoe} occupies a middle ground, employing autoregressive training on raw time series with a sparse mixture-of-experts (MoE) architecture to improve scalability and model capacity.

Beyond their training paradigms, TSFMs also differ substantially in architectural design and input representations. Moirai adopts an encoder-only architecture with patch-wise embeddings, which facilitates efficient representation learning and long-context modeling but may obscure fine-grained temporal variations. Chronos and Time-MoE instead rely on point-wise embeddings and autoregressive generation, allowing more precise modeling of local dynamics at the cost of increased sensitivity to input length and error accumulation in long-horizon forecasting. These design choices directly influence how TSFMs can be repurposed for downstream tasks such as anomaly detection, particularly in online or zero-shot settings where forecasting behavior is critical.

\subsection{Parameter-Efficient Fine-Tuning}
While TSFMs exhibit strong generalization, adapting them to domain-specific tasks often requires some form of fine-tuning. Parameter-efficient fine-tuning has emerged as an effective strategy for specializing large pretrained models while avoiding the computational and memory costs of full fine-tuning. Representative methods include IA$^3$ \cite{liu2022fewshot}, which introduces trainable scaling vectors to modulate key transformer activations, and LoRA \cite{hu2022lora}, which injects low-rank updates into pretrained weight matrices. Other approaches, such as OFT \cite{qiu2023controlling}, restrict adaptation to orthogonal transformations to preserve geometric structure, while HRA \cite{yuan2024bridging} extends this idea using products of Householder reflections \cite{householder1958unitary} to balance stability and expressiveness.

Originally developed for large language and vision models, PEFT techniques are particularly appealing for TSFMs, where model sizes can be substantial and retraining from scratch is often impractical. However, despite their growing adoption in forecasting and representation learning, the role of PEFT in time series anomaly detection remains largely unexplored, especially in comparison to full fine-tuning and zero-shot inference.

\subsection{TSFMs for Anomaly Detection}
Recent work has begun to investigate the applicability of TSFMs to anomaly-related tasks, though existing studies remain limited in scope. Shyalika \textit{et al.}~\cite{shyalika2024time} distinguish between anomaly prediction and anomaly detection, arguing that predicting anomalies ahead of time is a more challenging problem that requires modeling precursor or causal dynamics. Their primary focus is therefore on anomaly prediction, which is formulated as a supervised classification or forecasting-with-labels task. While the authors also report zero-shot anomaly detection results using a small set of TSFMs, the performance is highly dataset-dependent and lacks consistent advantages over simpler baselines. Moreover, fine-tuning experiments performed via full model updates yield only modest improvements, suggesting that naive adaptation of TSFMs may not be sufficient for robust detection performance.

A more closely related effort is the MOMENT framework \cite{goswami2024moment}, which demonstrates that a single pretrained time series model can be fine-tuned for multiple downstream tasks, including forecasting, imputation, classification, and anomaly detection. MOMENT represents an important early step in showing that TSFMs can be repurposed beyond forecasting, and it reports strong zero-shot performance as well as competitive results using linear probing for anomaly detection. However, its evaluation is limited to relatively small anomaly detection benchmarks, and the study does not investigate parameter-efficient fine-tuning strategies beyond linear layers.

In contrast to these preliminary explorations, our work provides a systematic and comprehensive study of repurposing TSFMs for time series anomaly detection. We focus explicitly on detection rather than prediction, evaluate across multiple large-scale TSAD benchmarks, and conduct a detailed comparison between zero-shot inference, full fine-tuning, and a range of PEFT methods. This broader empirical analysis clarifies when and how TSFMs can serve as effective anomaly detectors, and highlights the practical benefits of PEFT for adapting foundation models to real-world TSAD scenarios.

\section{Methods}
\subsection{Dataset}
For empirical evaluation, we use the TSB-AD-U benchmark \cite{liu2024the}, a comprehensive univariate time series anomaly detection suite designed to address limitations of prior benchmarks through improved dataset integrity, diversity, and evaluation reliability. TSB-AD-U comprises 23 carefully curated univariate datasets spanning diverse domains, anomaly types, sequence lengths, and statistical characteristics, offering substantially broader coverage than earlier collections \cite{paparrizos2022tsb}. This diversity makes TSB-AD-U particularly well suited for evaluating foundation models and PEFT strategies, as it reduces overfitting to specific anomaly patterns and better reflects real-world heterogeneity, thereby providing a robust and reliable testbed for benchmarking TSAD methods. Following the benchmark protocol, TSB-AD-U provides predefined training, tuning, and evaluation splits. In this study, we use the tuning set for performance evaluation. While this split was originally intended for hyperparameter selection, we adopt it as the evaluation set due to its substantially smaller size compared to the original evaluation set, enabling more efficient experimentation. All hyperparameters are fixed prior to evaluation, and the tuning set is used exclusively for reporting final results to ensure a fair and unbiased assessment.

\subsection{TSFM Backbones}
We evaluate a diverse suite of time series foundation model backbones that vary in architecture, scale, and pretraining strategy. Importantly, our use of these models differs across experimental settings. Zero-shot anomaly detection is conducted using a broad range of TSFMs to assess generalization without task-specific adaptation, whereas full fine-tuning and parameter-efficient fine-tuning are applied only to a selected subset of representative backbone models.

\paragraph{\textnormal{\textbf{Zero-shot inference.}}}
For zero-shot evaluation, we include representative models from three major TSFM families: Moirai \cite{woo2024unified}, Chronos \cite{ansari2024chronos}, and Time-MoE \cite{shi2025timemoe}, as summarized in Table~\ref{tab:zero-shot}. From the Moirai family, we evaluate both dense and mixture-of-experts variants, including Moirai and Moirai-MoE \cite{liu2025moiraimoe}. In particular, Moirai-MoE-small activates approximately 11M parameters per forward pass, corresponding to roughly 10\% of its total parameters, while larger variants increase overall model capacity through additional experts and retain sparse expert activation \cite{woo2024unified}.
From the Chronos family, we evaluate both T5-based and Bolt-based backbones. The Chronos-Bolt variants include bolt-tiny (9M parameters), bolt-small (48M), and bolt-base (205M), following the official Chronos model cards \cite{ansari2024chronos}. For T5-based Chronos models, parameter counts align with the underlying T5 architectures---t5-small (46M parameters), t5-base (200M), and t5-large (710M), which are adapted for time series forecasting through numerical tokenization and autoregressive decoding \cite{raffel2020exploring,ansari2024chronos}. In addition, we include Time-MoE-base (50M parameters) per forward pass as a representative autoregressive mixture-of-experts TSFM pretrained directly on large-scale time series data \cite{shi2025timemoe}. This broad selection enables a systematic analysis of how architectural design, model capacity, and sparsity affect zero-shot anomaly detection performance.

\paragraph{\textnormal{\textbf{Fine-tuning and parameter-efficient adaptation.}}}
To study the effects of task-specific adaptation, we focus full fine-tuning and PEFT experiments on three representative backbone models: Moirai-base, Chronos-t5-base, and Time-MoE-base. These models are selected to span different architectural paradigms, including encoder-only patch-based modeling (Moirai), autoregressive transformer forecasting via numerical tokenization (Chronos-t5), and sparse mixture-of-experts autoregressive modeling (Time-MoE), while remaining computationally feasible for repeated adaptation experiments. Restricting fine-tuning and PEFT to this subset allows for controlled comparisons between zero-shot inference, full model updates, and lightweight adaptation strategies, while avoiding confounding effects introduced by large disparities in model scale or training cost.

\subsection{Adaptation Strategies and Training Protocol}
We consider two adaptation regimes for repurposing TSFMs toward time series anomaly detection: full fine-tuning and PEFT. Full fine-tuning updates all parameters of the pretrained backbone model using task-specific training data, serving as an upper-bound baseline in terms of adaptation flexibility but incurring substantial computational and memory costs. For PEFT, we apply several lightweight adaptation strategies that modify only a small subset of parameters while keeping the pretrained backbone largely frozen. Specifically, we evaluate LoRA \cite{hu2022lora}, OFT \cite{qiu2023controlling}, and HRA \cite{yuan2024bridging}, which introduce trainable adapter modules into the transformer architecture. Following standard practice \cite{stickland2019bert,he2022towards}, these adapters are inserted into the key, query, value, and output projection layers of the attention blocks, while biases and dropout layers are excluded from adaptation. For these methods, we experiment with adapter ranks of 4, 8, 16, and 32 to examine the trade-off between parameter efficiency and detection performance. In addition, we evaluate IA$^3$ \cite{liu2022fewshot}, which adapts pretrained models by learning multiplicative scaling vectors rather than additive low-rank updates. For IA$^3$, trainable vectors are applied to the key and value projections as well as the feed-forward network layers, enabling lightweight modulation of internal activations with minimal parameter overhead.

Across all fine-tuning experiments, models are trained for a fixed 10 epochs using a context window of 150 time steps. Each model is optimized using the loss function employed in its original TSFM formulation, namely negative log-likelihood for Moirai, cross-entropy for Chronos, and Huber loss for Time-MoE. To account for sensitivity to optimization hyperparameters, we explore learning rates of $10^{-2}$, $10^{-3}$, $10^{-4}$, and $10^{-5}$, and report results corresponding to the configuration that achieves the best average performance across all datasets. This protocol allows for a fair and systematic comparison between full fine-tuning and different PEFT strategies under consistent training conditions.

\subsection{Evaluation Metrics}
For performance evaluation, we adopt four threshold-free metrics: AUC-ROC, AUC-PR, VUS-ROC, and VUS-PR \cite{liu2024the,boniol2025vus}. These metrics assess anomaly detection performance across all possible decision thresholds, enabling robust model comparison without requiring manual or dataset-specific threshold tuning. AUC-ROC and AUC-PR quantify the trade-offs between true positive and false positive rates, and between precision and recall, respectively, but rely on point-wise alignment between predicted anomalies and ground-truth labels, thereby penalizing even minor temporal shifts in detection.

To mitigate this limitation, we additionally report VUS-ROC and VUS-PR, which extend traditional AUC-based metrics by incorporating variable buffer zones around ground-truth anomaly points. This design allows the evaluation to assign partial credit when detections occur near, but not exactly at, the annotated anomaly locations, making the metrics more tolerant to small temporal misalignments that commonly arise in forecasting-based anomaly detection. However, this added flexibility comes at a higher computational cost: while standard AUC metrics can be computed efficiently by sweeping over all unique anomaly scores as candidate thresholds, VUS-based metrics require recomputing detection outcomes across multiple buffer sizes for each threshold. For all methods, anomaly scores are computed as the mean squared error (MSE) between predicted and true values at each time step, providing a consistent basis for metric computation.

\subsection{Baseline Anomaly Detection Methods}
To contextualize the performance of TSFMs, we compare against a diverse set of widely used time series anomaly detection baselines that do not rely on foundation-model pretraining. These baselines span forecasting-based, reconstruction-based, statistical, and classical machine learning approaches, providing a comprehensive reference across different modeling paradigms. Specifically, we include Anomaly Transformer \cite{xu2022anomaly}, LSTMAD \cite{malhotra2015long}, and TimesNet \cite{wu2023timesnet} as representative deep learning–based forecasting or sequence modeling methods, where anomalies are identified through prediction errors or learned temporal irregularities. Reconstruction-based neural baselines include AutoEncoder \cite{sakurada2014anomaly} and OmniAnomaly \cite{su2019robust}, which detect anomalies via reconstruction errors under the assumption that abnormal patterns cannot be faithfully reconstructed.

In addition to neural TSAD baselines, we evaluate classical and statistical baselines commonly used in TSAD benchmarks, including Sub-PCA \cite{aggarwal2017outlier}, Isolation Forest (IForest) \cite{liu2008isolation}, and SAND \cite{boniol2021sand}. These methods rely on dimensionality reduction, tree-based isolation, or statistical deviation modeling, and remain competitive in certain scenarios despite their simplicity. Together, this collection of baselines enables a fair and well-rounded comparison, highlighting the benefits and limitations of repurposing TSFMs relative to established anomaly detection techniques.

\begin{table}[t]
    \centering
    \caption{Performance comparison of baseline time series anomaly detection methods on the TSB-AD-U benchmark. Results are reported using threshold-free metrics (AUC-PR, AUC-ROC, VUS-PR, and VUS-ROC), averaged across all datasets. The baselines include traditional machine learning approaches, deep learning–based forecasting, and reconstruction models. Best and second-best results are highlighted in bold and underline, respectively.}
    \resizebox{0.8\textwidth}{!}{%
    \begin{tabular}{cccccc}
        \toprule
        Baseline & Model Size & AUC-PR & AUC-ROC & VUS-PR & VUS-ROC\\
        \midrule
        Sub-PCA  & - & 0.293 & 0.700 & \underline{0.332} & 0.745\\
        IForest  & - & 0.187 & 0.659 & 0.234 & 0.732\\
        SAND     & - & \underline{0.299} & \underline{0.714} & \textbf{0.353} & 0.747\\
        \midrule
        AutoEncoder & $\sim 7.7K$ & 0.173 & 0.667 & 0.242 & 0.729\\
        LSTMAD & $\sim 10K$ & \textbf{0.332} & \textbf{0.724} & 0.323 & \textbf{0.820}\\
        OmniAnomaly & $\sim 13K$ & 0.263 & 0.687 & 0.308 & 0.735\\
        TimesNet & $\sim 73K$ & 0.184 & 0.637 & 0.259 & \underline{0.748}\\ 
        Anomaly Transformer & $\sim 4.7M$ & 0.059 & 0.499 & 0.102 & 0.575\\
        \bottomrule
    \end{tabular}}
    \label{tab:baseline}
\end{table}

\section{Experimental Result}
We evaluate anomaly detection performance on the TSB-AD-U benchmark, which comprises 23 diverse univariate datasets spanning multiple domains and anomaly characteristics. Overall, the results in Tables \ref{tab:baseline}–\ref{tab:adaptation} demonstrate that TSFMs provide substantial advantages over task-specific anomaly detection baselines. In particular, TSFMs achieve strong zero-shot performance and can be further improved through both full fine-tuning and PEFT, confirming their effectiveness and flexibility for time series anomaly detection.

\begin{table}[t]
    \centering
    \caption{Zero-shot anomaly detection performance of time series foundation models on the TSB-AD-U benchmark. All models are evaluated without task-specific fine-tuning, using forecasting errors as anomaly scores. The table compares different TSFM families (Moirai, Chronos, and Time-MoE), model scales, and architectural variants, highlighting the impact of model design and capacity on zero-shot detection performance. Best and second-best results within each metric are shown in bold and underline, respectively.}
    \resizebox{0.7\textwidth}{!}{%
    \begin{tabular}{cccccc}
        \toprule
        TSFM & Model Size & AUC-PR & AUC-ROC & VUS-PR & VUS-ROC\\
        \midrule
         \multirow{6}{*}{Moirai}   
         & small & 0.335 & \textbf{0.697} & \underline{0.318} & \textbf{0.778}\\
         & base & 0.321 & 0.678 & 0.312 & 0.763\\
         & large & 0.326 & 0.692 & 0.315 & 0.771\\ \cmidrule(lr){2-6}
         & moe-small & 0.284 & 0.641 & 0.294 & 0.747\\
         & moe-base & 0.283 & 0.641 & 0.291 & 0.747\\
         \midrule
         \multirow{6}{*}{Chronos}   
         & t5-small & 0.304 & 0.675 & 0.280 &0.754\\
         & t5-base & 0.314 & 0.667 & 0.291 & 0.753\\
         & t5-large & 0.323 & 0.664 & 0.290 & 0.752 \\ \cmidrule(lr){2-6}
         & bolt-tiny& \underline{0.345} & \underline{0.693} & 0.318 & \underline{0.774}\\ 
         & bolt-small& \textbf{0.346} & 0.684 & \textbf{0.319} & 0.770\\
         & bolt-base & 0.340 & 0.678 & 0.317 & 0.765\\
         \midrule
         Time-MoE & base & 0.339 & 0.669 & 0.318 & 0.756\\
         \bottomrule
     \end{tabular}}
     \label{tab:zero-shot}
\end{table}

\subsection{Conventional TSAD Baselines}
Table \ref{tab:baseline} summarizes the performance of conventional TSAD baselines. Among classical and statistical approaches, SAND and Sub-PCA achieve the strongest overall results, especially under VUS-based metrics, indicating robustness to temporal misalignment. Among neural models trained from scratch, LSTMAD performs best across most metrics, achieving the highest AUC-PR, AUC-ROC, and VUS-ROC in this group. In contrast, more complex architectures such as anomaly transformer perform poorly on average, suggesting limited robustness across the heterogeneous datasets in TSB-AD-U.

\subsection{Comparison with Zero-Shot Performance of TSFMs}
Table~\ref{tab:zero-shot} reports the zero-shot anomaly detection performance of TSFMs on the TSB-AD-U benchmark, evaluated using forecasting errors without any task-specific adaptation. Overall, zero-shot TSFMs achieve competitive performance relative to established baseline methods shown in Table~\ref{tab:baseline}, but do not consistently surpass the strongest task-specific models across all metrics. In particular, while several TSFM variants attain AUC-PR values comparable to or slightly exceeding those of traditional and deep learning baselines, LSTMAD remains the top-performing method in terms of AUC-ROC and VUS-ROC.

Among TSFMs, models from the Chronos and Moirai families generally outperform other zero-shot configurations. Chronos Bolt variants achieve the highest AUC-PR and VUS-PR scores among TSFMs, while dense Moirai models exhibit strong AUC-ROC and VUS-ROC performance. However, the performance differences across TSFM architectures and scales are relatively modest, and mixture-of-experts variants do not consistently improve zero-shot results over their dense counterparts. These results suggest that, although directly applying pretrained TSFMs without adaptation can yield robust and competitive anomaly detection performance across diverse datasets, zero-shot inference alone is insufficient to consistently outperform well-tuned task-specific baselines. This observation motivates the need for task-specific adaptation strategies, which we explore in the subsequent section.

\begin{figure}[t]
    \centering
    \includegraphics[width=0.7\linewidth]{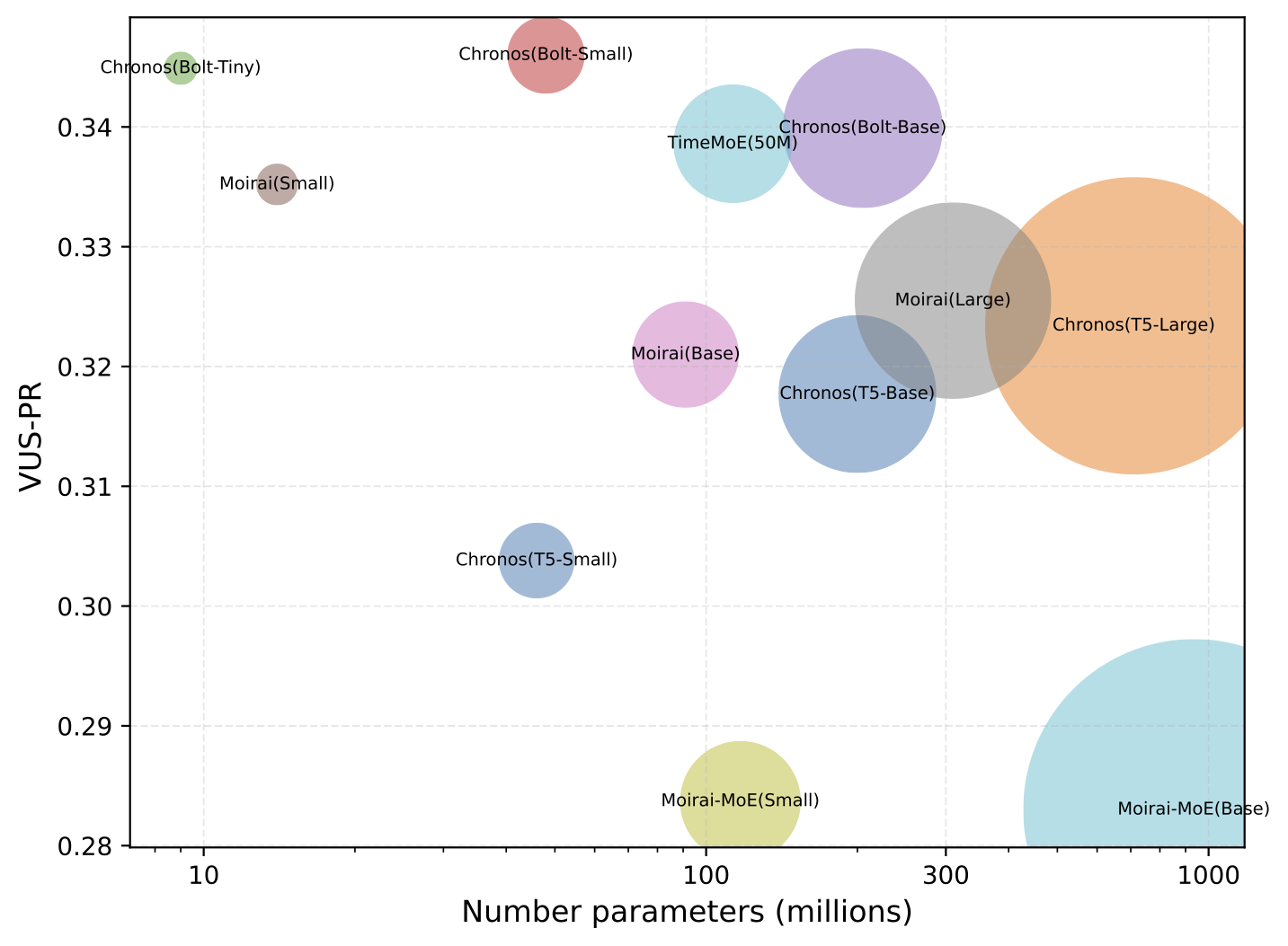}
    \caption{Relationship between model size and zero-shot anomaly detection performance. Mean VUS-PR is plotted against the number of model parameters for different TSFM variants evaluated on the TSB-AD-U benchmark. Bubble size reflects relative model capacity.}
    \label{fig:scale_vs_vuspr}
\end{figure}

\subsection{Impact of Adaptation Strategies on Performance}
A closer per-metric analysis in Table~\ref{tab:adaptation} shows that adapting TSFMs yields consistent improvements over zero-shot inference across all backbones, with gains being most pronounced for AUC-PR and VUS-PR. These metrics are particularly relevant under the severe class imbalance characteristic of anomaly detection. While improvements in AUC-ROC are generally more moderate for dense TSFMs, reflecting that zero-shot models already provide reasonable ranking performance, Time-MoE exhibits substantial gains under adaptation, indicating that mixture-of-experts architectures benefit more strongly from task-specific calibration. Overall, precision-oriented and VUS-based metrics exhibit clearer gains after adaptation than ranking-based metrics alone, suggesting that adaptation primarily improves the accurate identification of rare anomalies rather than merely refining global ranking behavior. In addition, the higher VUS-ROC and VUS-PR scores observed after adaptation indicate improved tolerance to small temporal misalignments, a desirable property when anomaly boundaries are ambiguous in practice. Architectural comparisons further show that while dense TSFMs are strong performers in the zero-shot setting, Time-MoE closes this gap and achieves competitive or superior VUS-based performance after adaptation. This suggests that explicit temporal and expert-level specialization, when properly fine-tuned, can outweigh the robustness advantages of dense architectures.

\begin{table}[t]
    \centering
    \caption{Comparison of adaptation strategies for repurposing TSFMs toward time series anomaly detection. Results are shown for zero-shot inference, full fine-tuning (Full-FT), and PEFT methods, including IA$^3$, LoRA, OFT, and HRA, across three representative backbone models. Performance is evaluated on the TSB-AD-U benchmark using threshold-free metrics. The table illustrates the trade-offs between adaptation flexibility and detection performance. Best and second-best results are highlighted in bold and underline, respectively.}
    \resizebox{0.8\textwidth}{!}{%
    \begin{tabular}{cccccc}
        \toprule
         TSFM & Adaptation &AUC-PR & AUC-ROC & VUS-PR & VUS-ROC\\
         \midrule
         \multirow{6}{*}{Moirai (base)} & Zero-shot & 0.321 & 0.678 & 0.312 & 0.763\\ \cmidrule(lr){2-6}
         & Full-FT& 0.387 & \underline{0.757} & 0.354 & 0.827\\ \cmidrule(lr){2-6}
         & IA$^3$& 0.377 & 0.748 & 0.351 & 0.817\\
         & LoRA & \underline{0.388} & \textbf{0.763} & 0.352 & \underline{0.829}\\
         & OFT & \textbf{0.389} & 0.756 & \textbf{0.361} & \textbf{0.834}\\
         & HRA & 0.387 & 0.751 & \underline{0.357} & 0.822\\
         \midrule
         \multirow{6}{*}{Chronos-t5 (base)} & Zero-shot & 0.318& 0.675 & 0.290 & 0.752\\ \cmidrule(lr){2-6}
         & Full-FT & \underline{0.361} & 0.727 & 0.318 & 0.796\\ \cmidrule(lr){2-6}
         & IA$^3$ & 0.325 & 0.680 & 0.299 & 0.765\\
         & LoRA & 0.325 & 0.674 & 0.297 & 0.757\\
         & OFT & \underline{0.361} & \underline{0.733} & \textbf{0.338} & \underline{0.813}\\
         & HRA & \textbf{0.367} & \textbf{0.738} & \underline{0.336} & \textbf{0.816}\\
         \midrule
         \multirow{6}{*}{Time-MoE (base)} & Zero-shot & 0.339 & 0.669 & 0.318 & 0.756\\ \cmidrule(lr){2-6}
         & Full-FT & \textbf{0.386} & 0.750 & \textbf{0.392} & \textbf{0.843}\\ \cmidrule(lr){2-6}
         & IA$^3$ & 0.347& 0.712& 0.341& 0.787\\
         & LoRA & \underline{0.384}& \underline{0.757}& \underline{0.372}& 0.830\\
         & OFT & 0.379& \textbf{0.762}& 0.365& \underline{0.840}\\
         & HRA & 0.351& 0.736& 0.351& 0.813\\
         \bottomrule
    \end{tabular}}
    \label{tab:adaptation}
\end{table}

\begin{figure}[t]
    \centering
    \includegraphics[width=1.0\linewidth]{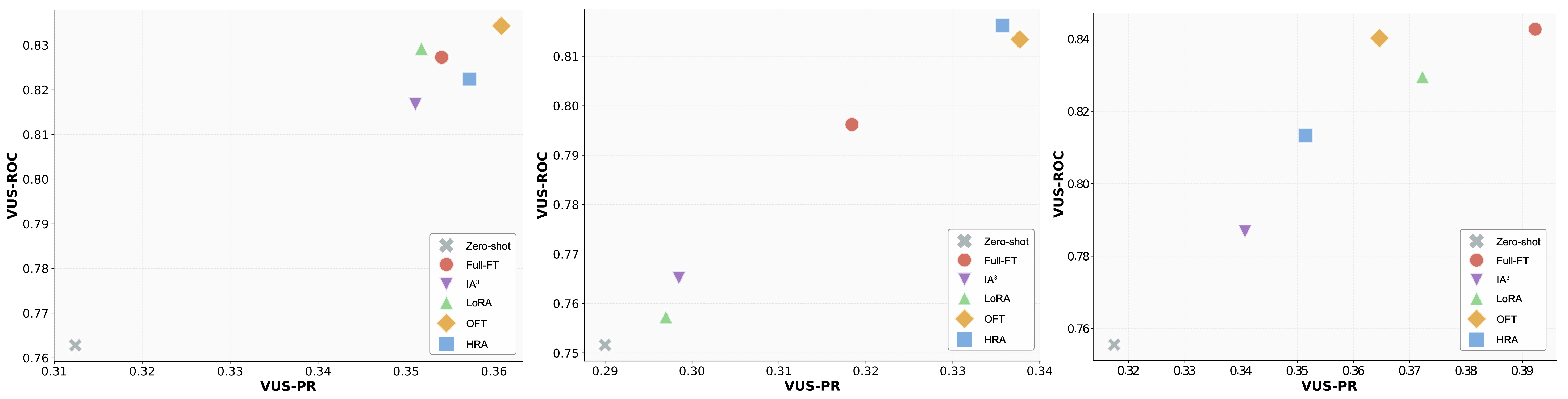}
    \caption{Comparison of VUS-PR and VUS-ROC performance for zero-shot, full fine-tuning, and PEFT methods on Moirai (left), Chronos (middle), and Time-MoE (right).}
    \label{fig:pr_vs_roc}
\end{figure}

Across all three backbone models, full fine-tuning consistently improves performance relative to zero-shot inference. In particular, Time-MoE shows the largest absolute gains on VUS-based metrics, with VUS-PR increasing from 0.318 to 0.392 and VUS-ROC from 0.756 to 0.843 after full fine-tuning. Moirai-base similarly improves its VUS-ROC from 0.763 to 0.827, while Chronos-t5 exhibits consistent gains across all evaluated metrics. These results confirm that adapting pretrained forecasting models to the target data distribution helps capture dataset-specific temporal patterns. However, the magnitude of improvement differs substantially across architectures, indicating that adaptation effectiveness depends not only on data alignment but also on how architectural components, such as expert routing or shared representations, interact with the adaptation strategy.

Parameter-efficient fine-tuning methods frequently match or surpass full fine-tuning while updating only a small fraction of model parameters (see Table~\ref{tab:peft_params}). For Moirai-base, adapter-based approaches such as LoRA, OFT, and HRA consistently outperform full fine-tuning across most metrics, with OFT achieving the best VUS-PR and VUS-ROC (Figure~\ref{fig:pr_vs_roc}). For Chronos-t5, OFT and HRA deliver the strongest overall results (Figure~\ref{fig:pr_vs_roc}), exceeding both zero-shot inference and full fine-tuning, particularly on precision-oriented and VUS-based metrics. In contrast, for Time-MoE, full fine-tuning achieves the strongest overall VUS-based performance, while LoRA and OFT only partially recover these gains (Figure~\ref{fig:pr_vs_roc}). This suggests that standard PEFT methods may be insufficient to fully adapt expert selection, load balancing, or inter-expert coordination mechanisms intrinsic to mixture-of-experts models. While PEFT still provides substantial improvements over zero-shot inference, the residual gap to full fine-tuning implies that Time-MoE may require more specialized PEFT designs such as expert-aware adapters or routing-aware parameterization to fully realize its adaptation potential.

\begin{table}[t]
    \centering
    \caption{Trainable parameters and ratios used in PEFT for Moirai, Chronos-t5, and Time-Moe.}
    \begin{tabular}{cccccll}
    \toprule
    & \multicolumn{2}{c}{Moirai (base)} & \multicolumn{2}{c}{Chronos-t5 (base)} &  \multicolumn{2}{c}{Time-MoE (base)}\\
    \cmidrule(lr){2-3} \cmidrule(lr){4-5} \cmidrule(lr){6-7}
    PEFT & \# params & ratio(\%) & \# params & ratio(\%)  & \# params & ratio(\%)\\
    \midrule
    IA$^3$ & $\sim 43K$ & 0.05 & $\sim 0.1M$ & 0.06 & $\sim 27K$ & 0.02\\
    LoRA   & $\sim 1.1M$ & 1.28 & $\sim 3.5M$ & 1.73 & $\sim 1.6M$ & 1.47\\
    OFT    & $\sim 0.8M$  & 0.94 & $\sim 2.5M$ & 1.27 & $\sim 4M$ & 3.48\\
    HRA    & $\sim 0.5M$  & 0.64 & $\sim 1.7M$ & 0.87 & $\sim 0.7M$ & 0.65\\
    \bottomrule
    \end{tabular}
    \label{tab:peft_params}
\end{table}

Finally, comparing PEFT variants, adapter-based methods (LoRA, OFT, and HRA) consistently outperform IA$^3$ across all backbones and evaluation metrics. Although IA$^3$ improves upon zero-shot performance in some cases, its gains are smaller and less consistent than those of adapter-based approaches, particularly for VUS-based metrics. This pattern suggests that multiplicative modulation alone lacks sufficient expressivity to recalibrate complex temporal and expert-level dynamics, whereas additive or orthogonal adapters provide a richer adaptation space. Consequently, adapter-based PEFT emerges as a more effective and reliable strategy for repurposing TSFMs for time series anomaly detection, especially when expert interactions are limited, as in dense architectures.

\section{Conclusion}
This work demonstrates the effectiveness of time series foundation models for anomaly detection, showing that even without task-specific training, TSFMs consistently outperform traditional baselines across a wide range of real-world datasets. By leveraging pretrained temporal knowledge acquired from large-scale and diverse time series corpora, TSFMs exhibit strong zero-shot performance and robust anomaly detection capability across domains. Moreover, parameter-efficient fine-tuning methods, including LoRA, OFT, and HRA, provide a compelling alternative to full fine-tuning, often matching or surpassing its performance while substantially reducing computational and memory overhead.

Despite these gains, important challenges remain, particularly in balancing architectural complexity, model size, and computational efficiency. While PEFT methods significantly enhance performance for both dense and mixture-of-experts TSFMs, further investigation is needed to better understand how adaptation strategies interact with different architectural designs at scale. Future work may explore more scalable PEFT variants, improved expert specialization for MoE-based TSFMs, and the extension of these models to broader anomaly detection settings, such as multivariate, streaming, or online scenarios. Overall, our findings highlight the strong potential of TSFMs combined with PEFT as a scalable and effective paradigm for time series anomaly detection in diverse real-world applications.

\subsubsection{Acknowledgements}
K.Y. is supported in part by the National Research Foundation of Korea (NRF) grant (No. RS-2024-00337092), the Institute of Information \& communications Technology Planning \& Evaluation (IITP) grants (No. RS-2020-II201373, Artificial Intelligence Graduate School Program; No. IITP-(2025)-RS-2023-00253914, Artificial Intelligence Semiconductor Support Program (Hanyang University)) funded by the Korean government (MSIT).

\bibliographystyle{splncs04}
\bibliography{reference} 

\end{document}